\documentclass[3p,11pt,authoryear]{elsarticle} 

\usepackage{lineno,hyperref}
\modulolinenumbers[1]
\usepackage{graphicx} % figures

\usepackage{float} 
\usepackage{subfigure}
\usepackage{color,soul}
\usepackage[version=4]{mhchem}
\usepackage{setspace}
\usepackage{booktabs} % table
\usepackage{siunitx} % SI units
\usepackage[section]{placeins} % make figures shown inside the cited sections
\usepackage{textcomp} % degree C
\usepackage[flushleft]{threeparttable} % table notes

\usepackage{array}
\newcolumntype{P}[1]{>{\raggedright\arraybackslash}p{#1}}

\usepackage{tabularx}
\usepackage{enumitem}
\newlist{tabitemize}{itemize}{1}
\setlist[tabitemize]{nosep,
                  topsep= 0pt,
                  partopsep=0pt,
                  leftmargin= *,
                  label=\textbullet,
                  before=\vspace{-0.6\baselineskip},
                  after=\vspace{-\baselineskip}
                  }
\usepackage{booktabs}

\usepackage[skip=3pt,format=plain,font=small,labelfont=bf]{caption}

\makeatletter
\def\ps@pprintTitle{%
 \let\@oddhead\@empty
 \let\@evenhead\@empty
 \def\@oddfoot{}%
 \let\@evenfoot\@oddfoot}
\makeatother

\makeatletter
\renewcommand\appendix{\par
  \setcounter{section}{0}%
  \setcounter{subsection}{0}%
  \setcounter{equation}{0}%
  \setcounter{table}{0}%------------ << add
  \setcounter{figure}{0}%----------- << add
  \gdef\theequation{\@Alph\c@section.\arabic{equation}}%
  \gdef\thefigure{\@Alph\c@section.\arabic{figure}}%
  \gdef\thetable{\@Alph\c@section.\arabic{table}}%
  \gdef\thesection{\appendixname\@Alph\c@section}%
  \@addtoreset{equation}{section}%
  \@addtoreset{table}{section}%----- << add
  \@addtoreset{figure}{section}%---- << add
}
\makeatother

\usepackage{etoolbox} % overlapping appendices
\makeatletter
\newrobustcmd{\fixappendix}{%
  \patchcmd{\l@section}{1.5em}{7em}{}{}%
  \patchcmd{\l@subsection}{2.3em}{7em}{}{}%
}
\makeatother

\begin{document}

\begin{frontmatter}

\title{\textbf{%BOOK CHAPTER \\ 
%\small{in \textit{Leveraging Artificial Intelligence in Engineering, Management, and Safety of Infrastructure}} \\
\mbox{Artificial Intelligence in Concrete Materials:~A Scientometric View}}}

\author[add1]{Zhanzhao Li\corref{cor1}}
\ead{zzl244@psu.edu}

\author[add1]{Aleksandra Radlińska}

\cortext[cor1]{Corresponding author}
\address[add1]{Department of Civil and Environmental Engineering, The Pennsylvania State University, University Park, PA 16802, USA}

\begin{abstract}
Artificial intelligence (AI) has emerged as a transformative and versatile tool, breaking new frontiers across scientific domains. Among its most promising applications, AI research is blossoming in concrete science and engineering, where it has offered new insights towards mixture design optimization and service life prediction of cementitious systems. This chapter aims to uncover the main research interests and knowledge structure of the existing literature on AI for concrete materials. To begin with, a total of 389 journal articles published from 1990 to 2020 were retrieved from the Web of Science. Scientometric tools such as keyword co-occurrence analysis and documentation co-citation analysis were adopted to quantify features and characteristics of the research field. The findings bring to light pressing questions in data-driven concrete research and suggest future opportunities for the concrete community to fully utilize the capabilities of AI techniques.
\end{abstract}

\begin{keyword}
	Artificial intelligence \sep Machine learning \sep Concrete materials \sep Scientometric analysis
\end{keyword}

\end{frontmatter}

%\linenumbers
%\newpage
%\tableofcontents
%\newpage

\section{Introduction}
\noindent
The world population has grown from 1.5 to 7.7 billion over the last hundred years \citep{Leridon2020}, with more than half of the humanity currently living in urban areas \citep{Chen2019a}. As the world population continues to grow, global urbanization will expand at a fast pace. It is expected that by 2050, two in every three people will live in cities \citep{Buhaug2013}. Such a high rate of urban development requires enormous quantities of materials for the construction of residential housing, commercial buildings, sanitation facilities, and other parts of the infrastructure. Concrete is the principal building material for the construction and infrastructure industries \citep{Mehta2014}. Despite decades of research and analysis, some scientific and engineering questions on concrete materials regarding mixture design optimization and service life prediction still remain unanswered \citep{Dolado2011,DeRousseau2018}. While most of the early concrete research relied on expert knowledge and intuition, trial-and-error experiments, or physical modeling, recent advances in data-driven techniques such as artificial intelligence (AI) can provide fresh perspectives to tackle the existing research questions \citep{Cai2020,Asteris2021,Young2019}.

During the past few years, AI has attracted worldwide attention as the ``fourth paradigm of science'' owing to the exponential growth in computing power, higher accessibility of data repositories, and more availability of data science tools \citep{Agrawal2016}. Compared with labor-intensive experiments or computationally expensive simulations, AI techniques take advantage of existing data, automatically learn patterns from them, and perform tasks without explicit instructions \citep{Batra2020}. Such data-driven techniques have offered an alternative route to accelerate concrete mixture design and optimization in a more effective manner \citep{Young2019,Gunasekera2020}: large concrete mixture datasets are constructed and fed into AI models; the models then screen and generate new mixtures with desired properties; and the best mixture can be identified and validated by experimental and computational tests, with the outcomes appended to the collected datasets and iteratively calibrating the models. Nevertheless, the interactions between AI and concrete are still in the nascent stages and the full power of AI in concrete research is far from being realized. Thus, an overview of current research progress and future opportunities will be helpful for the wider adoption of AI in the construction industry.

Existing AI review studies in the concrete domain \citep{BenChaabene2020,Rafiei2016,DeRousseau2018,Behnood2021,Nunez2021a} have made valuable contributions. Yet, they have some limitations. First, these studies frequently adopted a relatively narrow perspective, focusing on limited AI applications in concrete science. For example, Nunez et al.~\citeyearpar{Nunez2021a} examined different AI models for accurate prediction of compressive strength of concrete; Ben Chaabene et al.~\citeyearpar{BenChaabene2020} further extended the scope to AI applications in concrete mechanical property prediction; while DeRousseau et al.~\citeyearpar{DeRousseau2018} discussed several computational methods used to optimize concrete mixture design. Second, previous review studies largely depended on the manual review and appraisal, which may lead to subjective interpretation \citep{Martinez2019,Pan2021}. As such, these review studies do not offer a full picture of the current knowledge structure of AI research within the concrete domain. As the first attempt to fill this gap, this chapter is intended to present a quantitative evaluation of the knowledge patterns and capture the research interests and emerging trends of AI in concrete science. 

%other studies have extended the scope to general property prediction of concrete at fresh and hardened states (such as predictions of strength, flow, and slump) \citep{Rafiei2016} and discussed the research trends of concrete mixture design optimization \citep{DeRousseau2018}.

To this end, the present chapter utilized scientometric techniques to conduct quantitative analysis on the literature regarding AI in concrete materials. Scientometrics is a branch of informatics that quantifies features and characteristics of science and scientific research, and unravels emerging research trends and knowledge structures in the research domain \citep{Chen2012}. While manual review provides an insightful overview of the research field, scientometric tools take scientific literature as an input and generate interactive visualization and mapping in a more effective and efficient manner \citep{Borner2003}. The objectives of this chapter are to: (1) quantitatively analyze existing literature within the topic by scientometric techniques (i.e., keyword co-occurrence analysis and documentation co-citation analysis); (2) highlight hot research topics and applications of AI in concrete materials; and (3) identify research gaps for further adoption of AI in the construction industry. 

%This chapter presents a comprehensive literature survey, followed by the scientometric analysis (i.e., keyword co-occurrence analysis and documentation co-citation analysis).

\section{Literature survey} \label{sec:literature_survey}
\noindent
This chapter analyzes bibliographic data retrieved from the Web of Science (WoS) Core Collection database, which is one of the most established bibliographic data sources and includes peer-reviewed, high-quality scholarly journals published worldwide \citep{Visser2021}. A combination with other bibliographic databases (such as Google Scholar and Scopus) was not considered due to difficulties in checking duplicates of publications and dealing with different data formats from various databases. 

Keywords were selected based on related review studies \citep{Halilaj2018,Darko2020,Martinez2019,Pan2021} and iterative search in the WoS platform. As a result, a list of keywords for topics related to concrete science and AI was created, with the query string being: (``concrete'' OR ``cement'' OR ``cementitious'') AND (``artificial intelligence'' OR ``machine learning'' OR ``deep learning'' OR ``neural network*'' OR ``support vector machine*'' OR ``random forest'' OR ``decision tree*'' OR ``k-nearest'' OR ``k nearest'' OR ``k-nn'' OR ``knn'' OR ``k-mean*'' OR ``k mean*'' OR ``gaussian process*''). Note that the wildcard character * was used to capture plural forms, e.g., ``neural network*'' matches ``neural network'' and ``neural networks''. The keyword search was performed on topic terms from the title, abstract, author keywords, and keywords plus (automatically generated from the titles of cited articles by WoS) within a record. The search period was set to include the last 30 years, from January 1990 to December 2020, which is sufficient based on our preliminary search to represent the development of AI within the concrete domain. The language of the publications was limited to English, and the ``document type'' was set to ``article'' such that the evolution of the field can be represented by high-quality and original research \citep{Darko2020}. We further restricted the ``research area'' by the Boolean operation (``Construction \& Building Technology'' AND ``Materials Science'' NOT ``Computer Science''), as keywords like ``concrete'' are often used in other senses of the words within other subject areas, such as computer science.

%Searches for topic terms, rather than merely for author keywords, returned results from title, abstract, author keywords, keywords plus (automatically generated from the titles of cited articles by WoS) within a record, enhancing the power of searching by retrieving all the publications containing the selected keywords in these fields. 

As of October 2021, 440 documents were identified. A further manual refining process was conducted to evaluate the collected articles by reading the source title and abstract for inclusion/exclusion. The exclusion was mainly for publications that were: (1) review articles and book chapters; (2) unrelated to concrete science (e.g., focused on other construction materials like steel but cited concrete-related articles as references); and (3) unrelated to AI (e.g., applied traditional statistical models like linear regression models and only included AI as a recommendation for future research). This approach filtered down the number of publications from 440 to 389. Detailed information of the retrieved publications is available in Li et al. \citeyearpar{Li2022a}.

\begin{figure}[!htb]
	\centering
	\includegraphics[width=\textwidth]{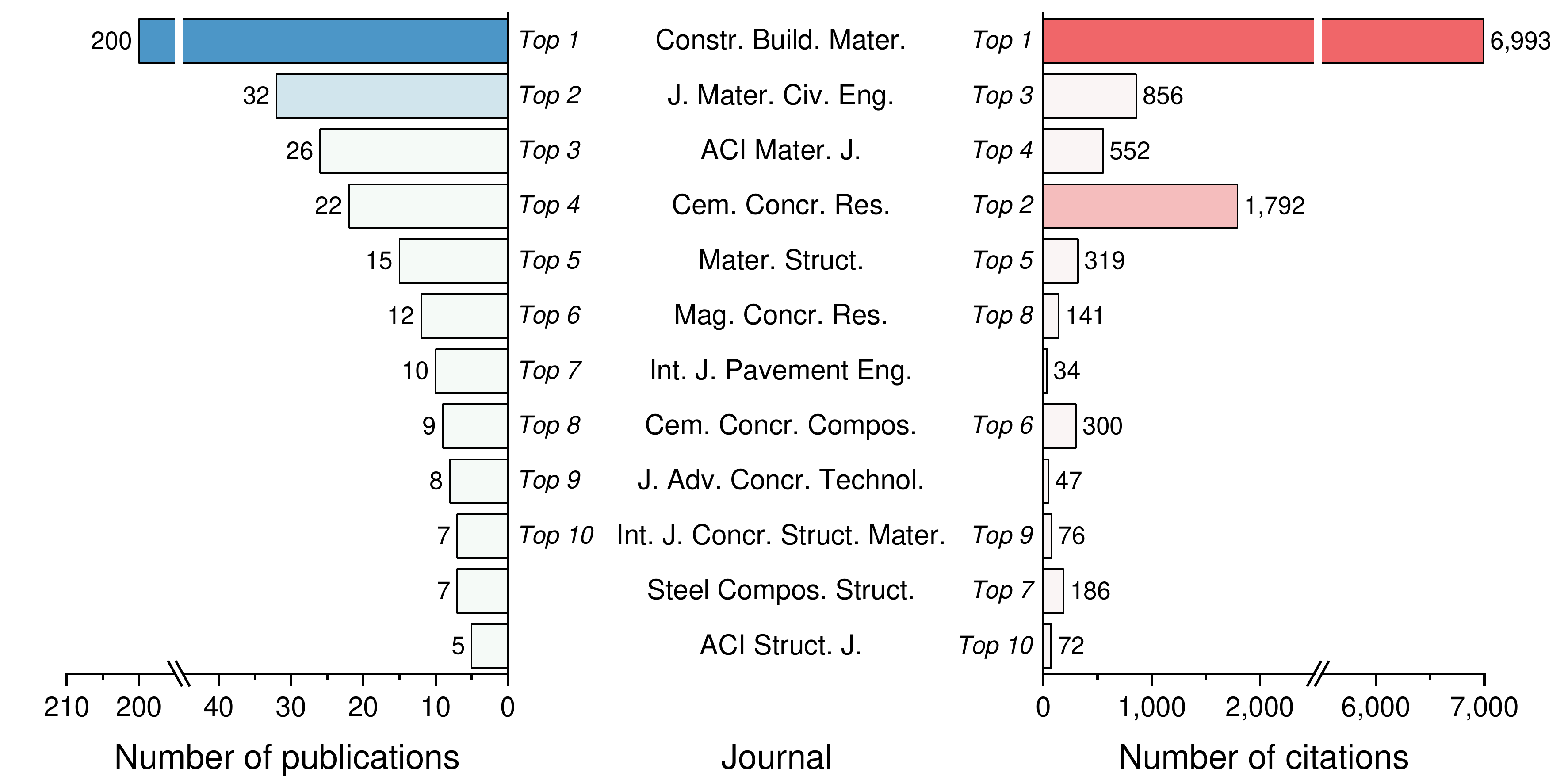}
	\caption{Top 10 journals in terms of numbers of publications and citations in the area of AI in concrete science. Journal titles are presented using abbreviations based on ISO (International Standards Organization) standards.}
	\label{fig:journal}
\end{figure}

Figure \ref{fig:journal} presents the list of top journals where publications on AI for concrete materials have been published. All the 12 journals in Fig.~\ref{fig:journal} had impact factors larger than 1.7 in the year 2020, which lends credence to the representativeness of the collected database. The majority (over 50\%) of the publications on AI in concrete have been published in \textit{Construction and Building Materials}, which is found to be the most influential journal in this research domain according to the number of citations (6,993). Interestingly, although the number of publications from \textit{Cement and Concrete Research} ranked fourth place, the citation record of this journal ranked second in its field. \textit{Steel and Composite Structures} and \textit{ACI Structural Journal} were found to be the main sources of references for citations, while they were not in the list of top 10 journals in terms of number of publications.

\begin{figure}[!htb]
	\centering
	\includegraphics[width=0.6\textwidth]{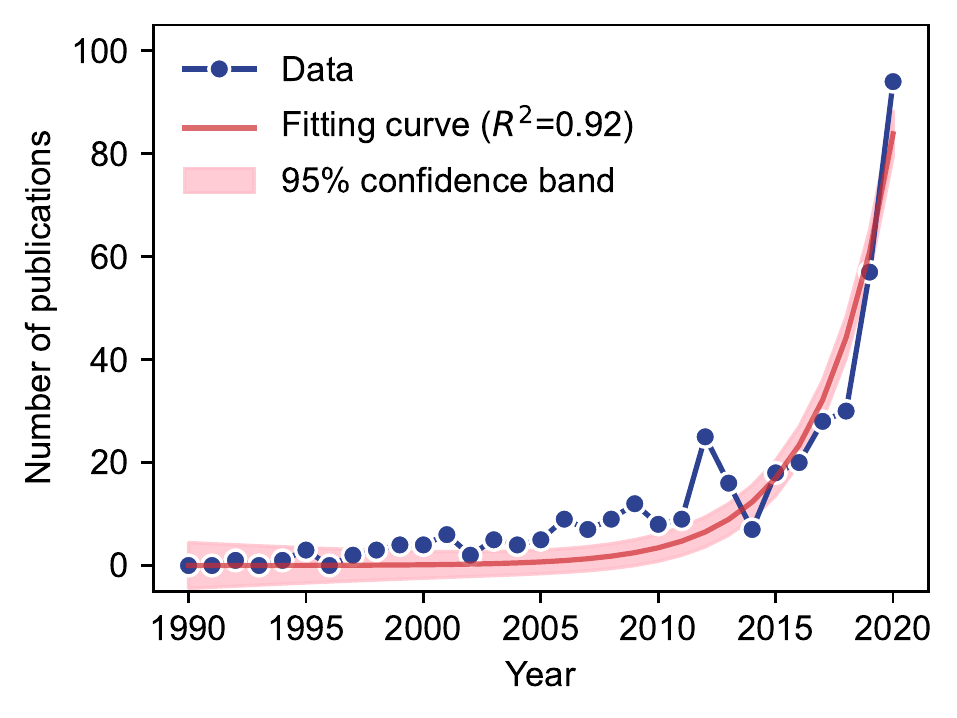}
	\caption{Number of publications for AI applications in concrete science during 1990--2020.}
	\label{fig:publication_year}
\end{figure}

The research field of AI dates back to the mid-20th century \citep{Haenlein2019} and has been extended to the construction industry in the 1970s \citep{Darko2020}; however, the adoption of AI in concrete materials has not started until the early 1990s. The first study on AI in the concrete domain appears to be the work from Pratt and Sansalone \citeyearpar{Pratt1992}, published in the \textit{ACI Materials Journal} in 1992, where AI was employed to automate signal interpretation for impact echo testing in the field. This was followed by the study from Mo and Lin \citeyearpar{Mo1994}, where they used AI to model the behavior of reinforced concrete framed shear walls. Figure \ref{fig:publication_year} shows the trend in research publications on AI in concrete science from 1990 to 2020. It reveals a relatively steady and gradual increase in research interest at the end of the 20th century. The number of publications did not reach double digits until 2009. Since then, exponential growth has been seen due to the ever-increasing computing power and availability of experimental data and computational tools. An exponential model was used to fit the data in Fig.~\ref{fig:publication_year}: $N(t)=N_{0}(1+p)^{\left(t-t_{0}\right)}$, where $N(t)$ and $N_{0}$ are the numbers of publications in the year of $t$ and the initial year $t_{0}=1990$, respectively. Parameter $p$ represents the annual percentage rate of increase. $N_{0}$ and $p$ were determined as 0.006 and 0.376, respectively. This indicates that the estimated increase in the publication number over the period of a year is 37.6\%, and that the numbers of relevant publications in 2021 and 2022 could be over 110 and 160, respectively. 

%That is to say, the application of AI in the concrete domain is gaining more and more attention under the innovations for digital concrete.
% Although some data points fall outside of the 95\% confidence band, the model gives a R-square of 0.92

\section{Main research interests: keyword co-occurrence analysis} \label{sec:keyword_co-occurrence_analysis}
\noindent
Keywords are the words or phrases that deliver the most essential content of a published document \citep{Martinez2019,Pan2021}. The analysis of keywords can offer opportunities to uncover the main research interests in any scientific field \citep{VanEck2014,Darko2020}. In order to construct and map the knowledge domain between AI and concrete materials, a keyword co-occurrence network was developed using VOSviewer \citep{VanEck2013}. The co-occurrence of two keywords can be defined as the situation when both keywords occur together in a publication \citep{VanEck2014}. A typical co-occurrence network of keywords consists of nodes (i.e., keywords) and links (indicating relations between pairs of nodes). The strength of the relation between two keywords represents the connection of their respective knowledge domains. Visualization of the keyword co-occurrence network provides an understanding of existing research topics and how they are intellectually developed and connected \citep{VanEck2014}. 

Keywords can be extracted from the title and abstract of a publication by text mining approaches or from the list of author-supplied keywords \citep{VanEck2014}. In this work, only author keywords were used to conduct the analysis and obtain reproducible and readable results. Identical terms (such as ``neural network'' and ``neural networks''; ``mix design'' and ``mixture design'') were merged for the accuracy of the analysis. However, keywords like ``concrete'', ``high performance concrete'', ``recycled aggregate concrete'', ``asphalt concrete'', ``self-consolidating concrete'', ``high strength concrete'', and ``reinforced concrete'' remained in the database, as their presence and the absence of the terms regarding other types of concrete indicated hot topics and research interests for the specific types of concrete \citep{Darko2020}. Other examples include the keywords ``strength'', ``compressive strength'', ``flexural strength'', and ``bond strength'', which are mechanical performance of hardened concrete. The minimum number of occurrences of a keyword was set to five such that 45 of the 969 keywords were included in the co-occurrence network. This criterion was selected following previous related studies \citep{Martinez2019} and based on multiple experiments to generate the optimal clusters in the network. Fractional counting methodology was adopted to fractionalize the weight of a link by considering the number of keywords in a publication \citep{VanEck2014}. This method has been found to be preferable over a traditional full counting approach \citep{Perianes-Rodriguez2016}. 

\begin{figure}[!htb]
	\centering
	\includegraphics[width=\textwidth]{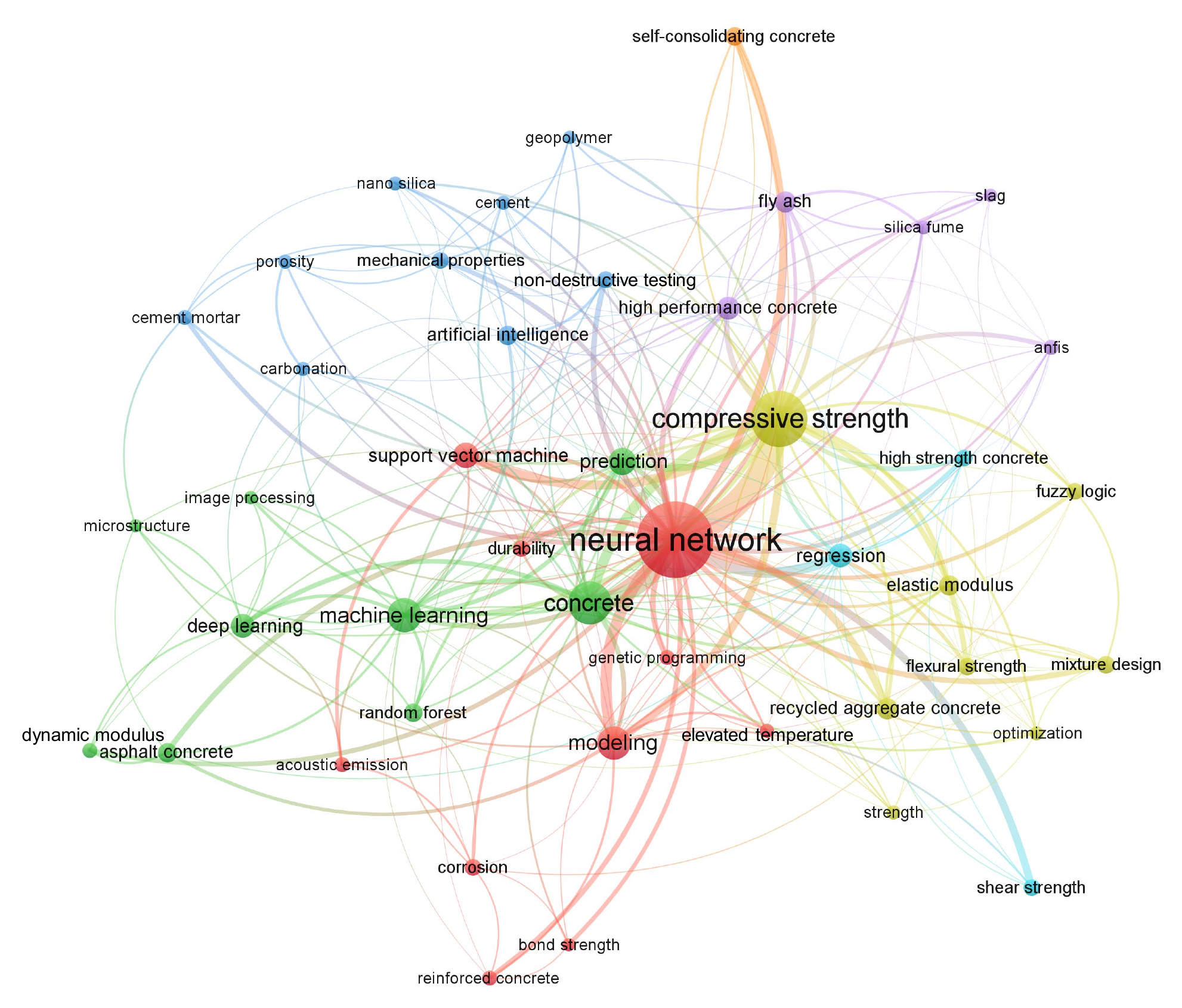}
	\caption{Network of co-occurring keywords related to research on AI in concrete materials (cluster view).}
	\label{fig:keyword_co-ocurrence}
\end{figure}

Figure \ref{fig:keyword_co-ocurrence} presents the resulting network of co-occurring keywords with 45 nodes and 536 links. In the network, each node represents a keyword term. The size of a node and a label denotes the number of publications that have the corresponding term in the keyword lists. The distance between two nodes indicates the relatedness of the nodes. In general, the smaller distance, the closer relationship between keywords. The width of a link represents the strength of the link, i.e., the number of publications where two keywords occur together. Nodes with the same color suggest a similar knowledge domain among the investigated publications \citep{VanEck2013}. Table \ref{tab:keywords} summarizes the keyword occurrences and related network data.  

\begin{table}[!htb]
%\footnotesize
\scriptsize
\centering
\caption{List of selected keywords and relevant network data (ranked by the number of occurrences).}
%\resizebox{\textwidth}{!}{%
\begin{threeparttable}
	\begin{tabularx}{\textwidth}{lllll}
		\toprule
		Keyword                     & Occurrences & Average publication year & Links & Total link strength \\ \midrule
		
		Neural network              & 167         & 2014                     & 42    & 130                 \\
		Compressive strength        & 92          & 2015                     & 32    & 85                  \\
		Concrete                    & 54          & 2014                     & 25    & 45                  \\
		Machine learning            & 34          & 2019                     & 26    & 29                  \\
		Modeling                    & 32          & 2011                     & 21    & 29                  \\
		Prediction                  & 22          & 2017                     & 23    & 22                  \\
		Support vector machine      & 19          & 2018                     & 11    & 17                  \\
		Deep learning               & 17          & 2019                     & 13    & 13                  \\
		Regression                  & 17          & 2015                     & 20    & 17                  \\
		High performance concrete   & 15          & 2015                     & 15    & 14                  \\
		Fly ash                     & 13          & 2014                     & 14    & 12                  \\
		Recycled aggregate concrete & 13          & 2016                     & 15    & 13                  \\
		Elastic modulus             & 12          & 2014                     & 15    & 11                  \\
		Artificial intelligence     & 11          & 2016                     & 15    & 9                   \\
		Asphalt concrete            & 11          & 2018                     & 6     & 9                   \\
		Random forest               & 10          & 2019                     & 7     & 7                   \\
		Self-consolidating concrete & 10          & 2013                     & 6     & 8                   \\
		Flexural strength           & 9           & 2018                     & 11    & 9                   \\
		Mixture design              & 9           & 2013                     & 11    & 9                   \\
		Corrosion                   & 8           & 2017                     & 9     & 6                   \\
		Durability                  & 8           & 2015                     & 14    & 8                   \\
		Fuzzy logic                 & 8           & 2011                     & 9     & 7                   \\
		High strength concrete      & 8           & 2013                     & 10    & 8                   \\
		Mechanical properties       & 8           & 2018                     & 11    & 8                   \\
		Non-destructive testing     & 8           & 2015                     & 9     & 8                   \\
		Shear strength              & 8           & 2013                     & 5     & 7                   \\
		Acoustic emission           & 7           & 2017                     & 5     & 6                   \\
		ANFIS                       & 7           & 2015                     & 12    & 7                   \\
		Cement mortar               & 7           & 2016                     & 6     & 7                   \\
		Dynamic modulus             & 7           & 2018                     & 5     & 6                   \\
		Genetic programming         & 7           & 2014                     & 7     & 7                   \\
		Reinforced concrete         & 7           & 2016                     & 6     & 6                   \\
		Carbonation                 & 6           & 2014                     & 10    & 6                   \\
		Cement                      & 6           & 2013                     & 10    & 6                   \\
		Elevated temperature        & 6           & 2017                     & 7     & 5                   \\
		Nano silica                 & 6           & 2017                     & 7     & 5                   \\
		Optimization                & 6           & 2014                     & 12    & 6                   \\
		Porosity                    & 6           & 2015                     & 7     & 6                   \\
		Silica fume                 & 6           & 2014                     & 9     & 6                   \\
		Strength                    & 6           & 2011                     & 9     & 4                   \\
		Bond strength               & 5           & 2015                     & 5     & 5                   \\
		Geopolymer                  & 5           & 2017                     & 5     & 5                   \\
		Image processing            & 5           & 2017                     & 6     & 5                   \\
		Microstructure              & 5           & 2015                     & 6     & 5                   \\
		Slag                        & 5           & 2011                     & 7     & 5                   \\ \bottomrule
	\end{tabularx} 
	\begin{tablenotes}
		\scriptsize
		\item Note: ANFIS = Adaptive neuro-fuzzy inference system.
	\end{tablenotes}	
\end{threeparttable} \label{tab:keywords}
\end{table}

As shown in Fig.~\ref{fig:keyword_co-ocurrence} and Table \ref{tab:keywords}, some research interests have received special attention. For example, keywords like ``modeling'', ``prediction'', ``regression'', and ``optimization'' have been frequently used in the publications. This could be explained by the fact that the main objectives of utilizing AI in concrete are to model concrete behavior, predict concrete properties, and optimize concrete mixture design for desirable performance. 

Interestingly, ``neural network'' appeared most frequently among all the keywords, indicating that neural network has been the most commonly used model, followed by support vector machine and random forest as shown in Table \ref{tab:keywords}. As the second most frequent keyword, ``compressive strength'' represented the strongest link with ``neural network'' (see Fig.~\ref{fig:keyword_co-ocurrence}). This shows that the most common AI application in concrete science has been the deployment of neural network models for compressive strength prediction of concrete, in agreement with other studies \citep{Behnood2021,BenChaabene2020}. Compressive strength is considered to be one of the most important design parameters in construction applications, since it determines the loading capacity of concrete structures and is correlated with several mechanical and durability properties of concrete, including tensile and flexural strength, elastic modulus, and impermeability \citep{ACICommittee2112002}. The prediction of compressive strength of concrete as a function of mixture proportions has become a research focus to facilitate mixture design optimization during the design phase and assist project scheduling as well as quality control during the production phase. Although AI techniques are promising, Table \ref{tab:keywords} suggests that other models such as fuzzy logic, adaptive neuro-fuzzy inference system (ANFIS), and genetic programming have obtained far less attention in concrete research. Thus, future research in this area should explore different AI models (i.e., models other than neural networks) and full development and exploitation of various models could be a promising aid in accelerating adoption of AI in the construction industry.

Figure \ref{fig:keyword_co-ocurrence} and Table \ref{tab:keywords} also show that research interests have been focused on the prediction of hardened properties for concrete materials, including compressive strength, elastic modulus, flexural strength, shear strength, bond strength, dynamic modulus, porosity, and durability (e.g., corrosion and carbonation). However, modeling of the performance of concrete in fresh state has remained understudied by AI techniques. This must draw attention of concrete researchers, given that fresh properties, such as setting, bleeding, segregation, heat evolution, plastic shrinkage, and rheological properties, are also of significance for both workability in construction practice and microstructural development towards the hardened performance of concrete (e.g., strength and durability) \citep{Kovler2011}. One major challenge that hinders practical applications of AI in fresh property prediction is the lack of large and universal datasets for concrete fresh properties. For example, absolute values (e.g., yield stress and plastic viscosity) calculated for a given concrete mixture are not identical and comparable among various rheometers (with different principles and geometries) \citep{Brower2003}. As such, data collection from different laboratories using different rheometers could become less meaningful. Moving forward, it would be promising to standardize reference materials for concrete rheometry in order to calibrate different concrete rheometers and normalize calculated values \citep{Ferraris2014}. With the development of large datasets for fresh properties and the availability of AI techniques, accurate prediction and modeling of concrete behavior at early ages could be very useful.

Based on the network, similar observations can be made for concrete types. Among various concrete types, high performance concrete, recycled aggregate concrete, asphalt concrete, self-consolidating concrete, high strength concrete, reinforced concrete, and geopolymer concrete have been extensively reported with AI methods (see Fig.~\ref{fig:keyword_co-ocurrence} and Table \ref{tab:keywords}). The absence of the other types of concrete in the network, including pervious concrete, lightweight aggregate concrete, and 3D-printed concrete, indicates that these types of concrete have been overlooked in AI-related concrete research. AI technologies have demonstrated their vast potential to offer novel approaches to model and predict material behaviors of the majority of concrete. Current research could be extended to explore other concrete types with the use of AI approaches, which could facilitate the development of an intelligence ecosystem for various concrete materials. 

As shown in Fig.~\ref{fig:keyword_co-ocurrence}, nodes with the same color suggest a similar topic among the publications, and seven distinct clusters of keywords were obtained in the network. From each cluster, main research interests and research gaps could be identified. For example, in the yellow cluster (Fig.~\ref{fig:keyword_co-ocurrence}, middle right), keywords such as ``strength'', ``compressive strength'', ``flexural strength'', ``elastic modulus'', were grouped together with ``mixture design'' and ``optimization''. This indicates that physical concrete performance measures have been the main objective for mixture design optimization \citep{Han2020,Ghafari2015a}. However, concrete mixture design involves multiple competing criteria, such as minimizing both cost and environmental impacts (e.g., carbon emissions) while maximizing physical performance \citep{Huang2020,Young2019,DeRousseau2018}. From the viewpoint of engineering applications, more research efforts would be needed to explore the application of AI techniques in multi-objective problems towards cheaper, stronger, more workable, durable, and more environmentally sustainable concrete.

\begin{figure}[!htb]
	\centering
	\includegraphics[width=\textwidth]{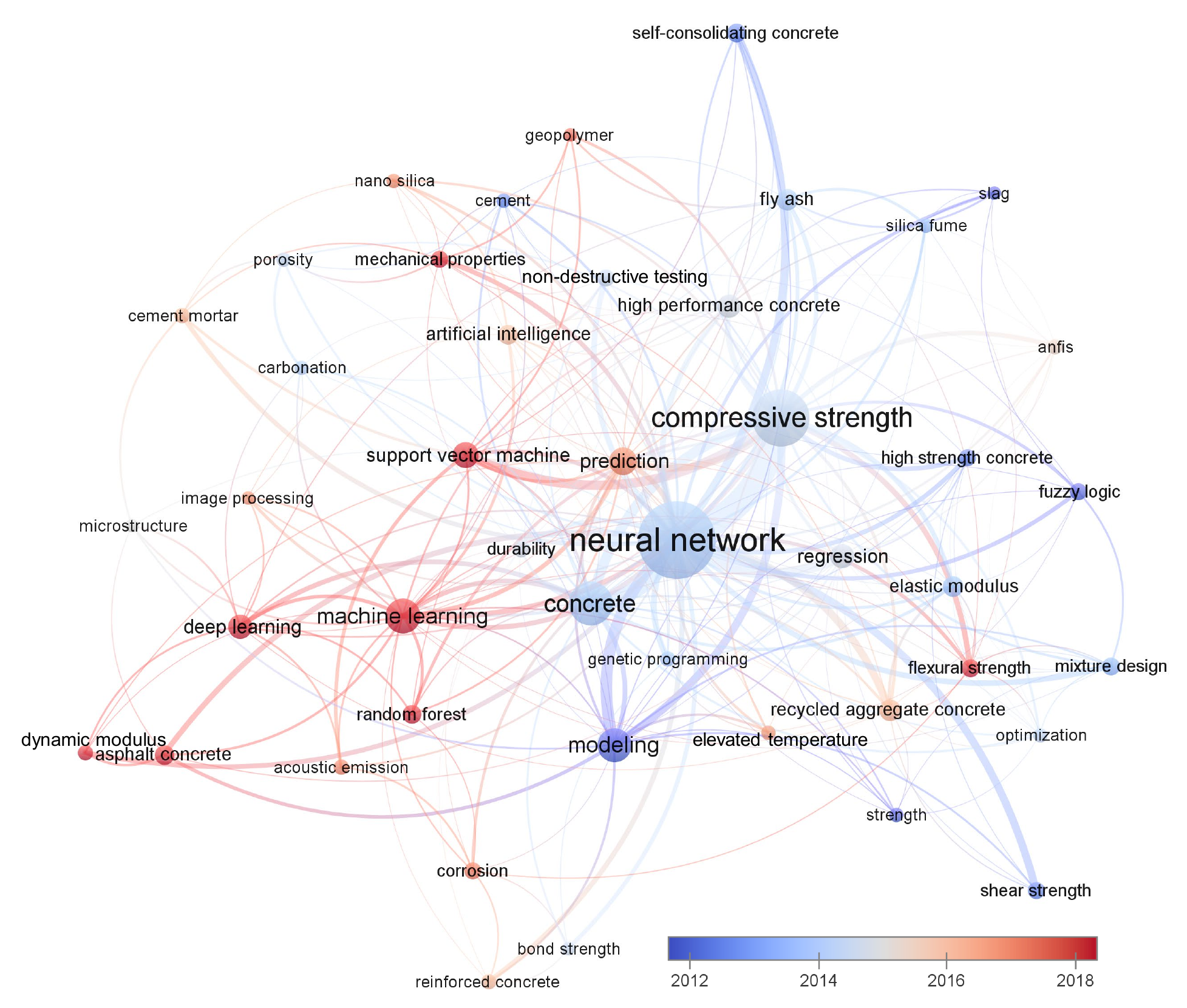}
	\caption{Network of co-occurring keywords related to research on AI in concrete materials (timeline view). Timeline covers a period of time from 2012 to 2018 in terms of average publication year.}
	\label{fig:keyword_co-ocurrence_timeline}
\end{figure}

The green cluster was another cluster with significant size in Fig.~\ref{fig:keyword_co-ocurrence}. This cluster consisted of keywords such as ``machine learning'', ``deep learning'', and ``image processing''. It is worth pointing out that these topics are considered emerging, as indicated in Fig.~\ref{fig:keyword_co-ocurrence_timeline}. In contrast to the static representations of knowledge domains in Fig.~\ref{fig:keyword_co-ocurrence}, the network in Fig.~\ref{fig:keyword_co-ocurrence_timeline} provides a timeline view of the knowledge map, where the evolution of AI applications in the concrete domain can be visualized according to the average publication year of each keyword. Notably, the average publication year of keyword ``deep learning'' was 2019, suggesting that research emphasis has been shifting towards the applications of deep learning techniques in concrete science, especially for image analysis and data recognition (corresponding to the keywords ``image processing'' and ``acoustic emission'', respectively). This could be attributed to the successful deep learning architectures developed in computer science, including convolutional neural networks for image classification \citep{Krizhevsky2012} and long short-term memory networks for time-series modeling \citep{Greff2017}. Note that deep learning techniques have been mainly used for crack detection in field practice \citep{Khallaf2021,Liu2019,Flah2020}, whereas the integration of these techniques with concrete petrographic analyses (e.g., flatbed scanning \citep{Song2020}, scanning electron microscopy \citep{Tong2019a,Tong2020}, and X-ray computed tomography \citep{Tong2019,Lorenzoni2020}) could be promising yet under-researched.

Keywords such as ``neural network'', ``compressive strength'', and ``concrete'' were represented in the middle spectrum around 2015 (see Fig.~\ref{fig:keyword_co-ocurrence_timeline} and Table \ref{tab:keywords}). This is because these topics have been the research focuses during the whole period of time investigated. AI models including support vector machine (average publication year: 2018) and random forest (average publication year: 2019) have become prevalent in addition to deep learning approaches, which indicates that more research efforts have been made recently to explore the effectiveness of AI techniques (other than neural networks) in concrete research.

\section{Citation patterns: document co-citation analysis} \label{sec:document_co-citation_analysis}
\noindent
A publication typically includes a list of references, which delineates the knowledge base of the publication. Similar to co-occurrence, co-citation could be defined as the frequency with which two references are cited together by other publications \citep{Small1973}. Given that frequently cited references typically present key concepts, methods, or conclusions in a research field, co-citation patterns of references could offer valuable insights into the relationship between these key ideas. The analysis of document co-citation has demonstrated its potential to unravel the intellectual structures of scientific literature and thus to identify research interests and emerging trends in a research domain \citep{Hou2018}.

%Hence, citation patterns of references in publications offer valuable insights regarding the intellectual structures and evolution of a research field \citep{Chen2012,Chen2014}. 
% The analysis of document co-citation examines the relationships between specific references and thus identifies research interests and emerging trends in scientific literature \citep{Hou2018}.
%facilitate the detection of emerging trends and abrupt changes in scientific literature \citep{Chen2012}
%Co-citation patterns change as the interests and intellectual patterns of the field change \citep{Small1973}. 

\begin{figure}[!htb]
	\centering
	\includegraphics[width=\textwidth]{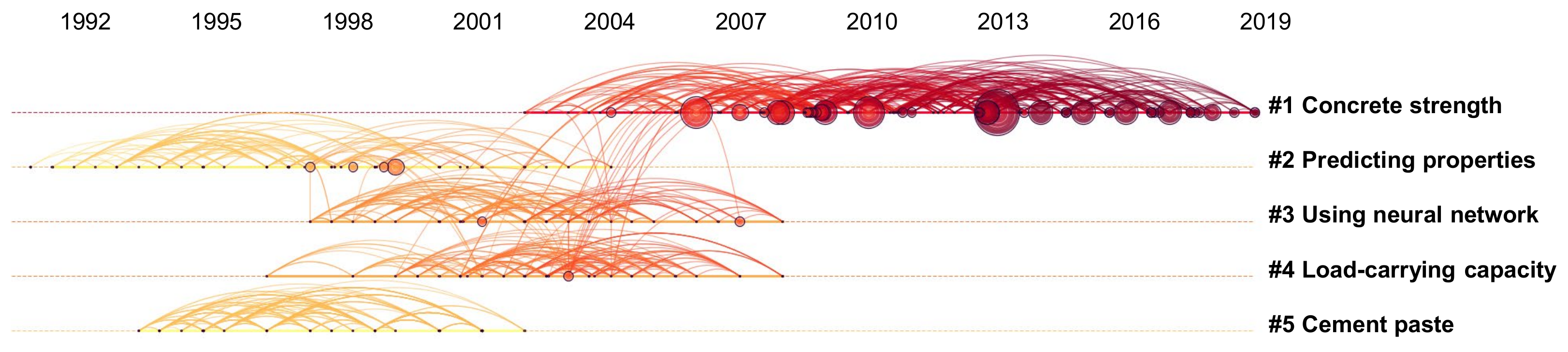}
	\caption{Network of co-cited documents in the publications on AI in concrete materials (timeline view).}
	\label{fig:document_co-citation_timeline}
\end{figure}

To further identify research trends in AI and concrete studies, a network of document co-citation was generated by CiteSpace \citep{Chen2014,Chen2016} (see Fig.~\ref{fig:document_co-citation_timeline}). Records of 389 citing publications and 11,842 cited references were retrieved from WoS, as described in Section \ref{sec:literature_survey}. In Fig.~\ref{fig:document_co-citation_timeline}, each node stands for a cited reference, and node size reflects the citation number of the reference. Links between nodes indicate co-citation. The colors of these links represent the first time when two references were co-cited. Links in more recent years are shown in red, whereas older co-citation links are colored in yellow. No network pruning (or link reduction) was performed to avoid reducing the characteristics of the bibliographic network. Figure \ref{fig:document_co-citation_timeline} and Table \ref{tab:document_co-citation} show the five major clusters identified in the network. These clusters were numbered and ranked in terms of their size, i.e., the number of references in a cluster. Each cluster was labeled using the log-likelihood ratio algorithm. This algorithm identifies a label based on the latent semantic analysis of publications that cite the references in each cluster. Although this approach has been found to provide the most accurate results in terms of unique labeling and sufficient coverage \citep{Chen2014}, additional manual examination of the publications was conducted to cross-validate the identified labels and determine the focus of each cluster.

\begin{table}[!htb]
%\footnotesize
\scriptsize
\centering
\caption{Major clusters identified in the document co-citation analysis.}
%\resizebox{\textwidth}{!}{%
	\begin{tabularx}{\textwidth}{lllp{2cm}ll}
		\toprule
		Cluster ID & Size & Silhouette value & Average   publication year & Cluster label            & Focus of the cluster                                     \\ \midrule
		
	1          & 169  & 0.913            & 2011                       & Concrete   strength      & Concrete compressive strength                 \\
	2          & 65   & 0.957            & 1996                       & Predicting   properties  & Mixture design         \\
	3          & 47   & 0.979            & 2002                       & Using neural network   & Neural network; durability               \\
	4          & 47   & 0.952            & 2002                       & Load-carrying capacity & Shear capacity; elastic modulus     \\
	5          & 46   & 0.976            & 1997                       & Cement paste           & Cement paste; waste solidification \\ \bottomrule
	\end{tabularx} \label{tab:document_co-citation}
\end{table}

Two fundamental metrics, namely, the modularity Q index and the silhouette value, were calculated to examine the reliability of the network. The modularity Q index ranges from 0 to 1 and assesses the extent to which a network can be partitioned into multiple clusters; the silhouette value ranges between --1 and 1 and measures the quality of a clustering configuration. A higher modularity Q index and a higher silhouette value imply higher divisibility of a network and better homogeneity of a cluster, respectively \citep{Lim2021,Chen2014}. The threshold of the silhouette value has been found to be 0.7, above which the clusters in a network could be considered highly isolated with clear boundaries and weak links between individual clusters \citep{Chen2014}. The network in Fig.~\ref{fig:document_co-citation_timeline} had a modularity Q index of 0.8867 and a weighted mean silhouette value of 0.9557. These values were relatively high, indicating that the network was highly divisible into clusters, each of which was, on average, highly homogeneous \citep{Chen2014}. In other words, studies in each identified cluster were consistent in addressing similar issues.

%The current body of knowledge on AI in the concrete domain comprises five major clusters (see Fig.~\ref{fig:document_co-citation_timeline} and Table \ref{tab:document_co-citation}). Each cluster showed a high silhouette value (i.e., over 0.9), which means that the five clusters identified in the network have a high extent of inner coherence.  

As per cluster sizes in Table \ref{tab:document_co-citation}, cluster \#1 on concrete strength has been the largest cluster with 169 references. After examining the content of the 47 publications that cited these references, the focus of this cluster could be further specified as concrete compressive strength \citep{Chou2014,Sobhani2010,Gholampour2017}. This is in line with the earlier observation in Fig.~\ref{fig:keyword_co-ocurrence} and Table \ref{tab:keywords}, where ``compressive strength'' has been the second most frequently used keyword. The ``average publication year'' in Table \ref{tab:document_co-citation} shows the average period within which the references in a given cluster were published. While most of the clusters showed activities over two decades ago (i.e., 1996--2002), studies in cluster \#1 have been published more recently, around 2011 on average, and this cluster had the longest active duration of approximately 17 years from 2002 to 2019. These again confirm the significance of cluster \#1 on concrete compressive strength in the field of AI in concrete science. 

Detection of citation burstness was further conducted to identify references that were of great significance to the concrete domain. Citation burstness occurs when a reference received a considerable surge in terms of citations and attention within a specific duration \citep{Chen2010} (e.g., large nodes in Fig.~\ref{fig:document_co-citation_timeline}). Table \ref{tab:top10_burstness} shows the top 10 references with the highest burst strengths (i.e., most co-cited and influential references). It is interesting to note that all these references listed in Table \ref{tab:top10_burstness} were grouped in cluster \#1. Most of them focused on the numerical prediction of compressive strength, and only two references \citep{Behnood2015,Duan2013a} on that of elastic modulus (which is correlated to compressive strength). All these references utilized AI techniques for prediction. In agreement with Section \ref{sec:keyword_co-occurrence_analysis}, neural network was the most commonly used algorithm \citep{Duan2013,Oztas2006,Sobhani2010,Pala2007,Topcu2008,Chithra2016,Dantas2013,Duan2013a}. The concrete type investigated in these studies varied from normal concrete \citep{Behnood2017,Pala2007,Topcu2008} to high performance concrete \citep{Behnood2017,Chithra2016} and recycled aggregate concrete \citep{Duan2013,Behnood2015,Duan2013a}. Note that most of these 10 cited references were included in the bibliographic database containing 389 citing publications, which cross-validates the completeness and representativeness of the retrieved database. Table \ref{tab:top10_most_cited} provides the citation information of the 389 publications regarding AI in concrete. Three studies \citep{Oztas2006,Duan2013,Sobhani2010} were present in both Table \ref{tab:top10_burstness} and Table \ref{tab:top10_most_cited}, suggesting their importance in the field of concrete research.

\begin{table}[!htb]
%\footnotesize
\scriptsize
\centering
\caption{Top 10 references regarding AI in concrete materials with the strongest document co-citation burstness.}
%\resizebox{\textwidth}{!}{%
	\begin{tabularx}{\textwidth}{lllllllll}
		\toprule
		No. & Reference                                                & Burst strength & Year & Begin & End  & Span & Citation counts & Cluster ID \\ \midrule
		
		1   & Duan et al.~\citeyearpar{Duan2013}          & 11.27          & 2013 & 2015  & 2020 & 5    & 29              & 1          \\
		2   & {\"{O}}ztaş et al.~\citeyearpar{Oztas2006}        & 10.15          & 2006 & 2008  & 2013 & 5    & 24              & 1          \\
		3   & Sobhani et al.~\citeyearpar{Sobhani2010}    & 7.84           & 2010 & 2012  & 2018 & 6    & 22              & 1          \\
		4   & Behnood et al.~\citeyearpar{Behnood2017}    & 7.76           & 2017 & 2017  & 2020 & 3    & 15              & 1          \\
		5   & Pala et al.~\citeyearpar{Pala2007}          & 7.66           & 2007 & 2011  & 2016 & 5    & 16              & 1          \\
		6   & Behnood et al.~\citeyearpar{Behnood2015}    & 7.59           & 2015 & 2018  & 2020 & 2    & 18              & 1          \\
		7   & Top{\c{c}}u and Saridemir \citeyearpar{Topcu2008} & 7.38           & 2008 & 2016  & 2020 & 4    & 16              & 1          \\
		8   & Chithra et al.~\citeyearpar{Chithra2016}    & 7.31           & 2016 & 2017  & 2020 & 3    & 16              & 1          \\
		9   & Dantas et al.~\citeyearpar{Dantas2013}      & 7.07           & 2013 & 2008  & 2013 & 5    & 17              & 1          \\
		10  & Duan et al.~\citeyearpar{Duan2013a}         & 6.12           & 2013 & 2015  & 2020 & 5    & 15              & 1         \\ \bottomrule
	\end{tabularx} \label{tab:top10_burstness}
\end{table}

\begin{table}[!htb]
%\footnotesize
\scriptsize
\centering
\caption{Top 10 most cited publications regarding AI in concrete materials.}
%\resizebox{\textwidth}{!}{%
	\begin{tabularx}{\textwidth}{lP{3cm}p{0.5cm}p{8.5cm}l}
		\toprule
		No. & Reference                                                        & Year & Title                                                                                                                             & Total citations \\ \midrule
		
		1   & Yeh \citeyearpar{Yeh1998}                              & 1998             & Modeling of strength of high-performance concrete using artificial neural   networks                                              & 548             \\
		2   & Gopalakrishnan et al.~\citeyearpar{Gopalakrishnan2017} & 2017             & Deep Convolutional Neural Networks with transfer learning for computer   vision-based data-driven pavement distress detection     & 254             \\
		3   & Ni and Wang \citeyearpar{Ni2000}                       & 2000             & Prediction of compressive strength of concrete by neural networks                                                                 & 243             \\
		4   & {\"{O}}ztaş et al.~\citeyearpar{Oztas2006}                   & 2006             & Predicting the compressive strength and slump of high strength concrete   using neural network                                    & 212             \\
		5   & Yeh \citeyearpar{Yeh2007}                              & 2007             & Modeling slump flow of concrete using second-order regressions and   artificial neural networks                                   & 163             \\
		6   & Duan et al.~\citeyearpar{Duan2013}                     & 2013             & Prediction of compressive strength of recycled aggregate concrete using   artificial neural networks                              & 152             \\
		7   & Sobhani et al.~\citeyearpar{Sobhani2010}               & 2010             & Prediction of the compressive strength of no-slump concrete: A   comparative study of regression, neural network and ANFIS models & 152             \\
		8   & Dorafshan et al.~\citeyearpar{Dorafshan2018}           & 2018             & Comparison of deep convolutional neural networks and edge detectors for   image-based crack detection in concrete                 & 150             \\
		9   & Alshihri et al.~\citeyearpar{Alshihri2009}             & 2009             & Neural networks for predicting compressive strength of structural light   weight concrete                                         & 138             \\
		10  & Tam et al.~\citeyearpar{Tam2007}                       & 2007             & Optimization on proportion for recycled aggregate in concrete using   two-stage mixing approach                                   & 135            \\ \bottomrule
	\end{tabularx} \label{tab:top10_most_cited}
\end{table}

As shown in Table \ref{tab:document_co-citation}, cluster \#2 with 65 cited references was identified as ``predicting properties''. There were eight citing publications in this cluster with a focus on mixture design \citep{Dias2001,Yeh1998,Yeh2006b,Yeh1998a,Yeh2007a}. Most studies \citep{Dias2001,Yeh2006b,Yeh1998} have emphasized sensitivity analyses for strength to investigate the effects of mixture constituents, such as cement, water, silica fume, and fly ash. Such sensitivity analyses could identify the relationship between strength and individual constituents of concrete mixtures. However, as mentioned earlier, there has been an ever-growing need for concrete mixtures to meet multiple objectives and satisfy many design criteria. Thus, developing appropriate AI models for multi-objective optimization (e.g., strength-cost-\ce{CO2} optimization) will contribute to the concrete industry \citep{Yeh1998a,Yeh2007a,Young2019}.

Cluster \#3 was automatically labeled as ``using neural network''. Manual examination of the five citing publications \citep{Oztas2006,Kwon2010,Pala2007,Chang2006,Song2009} confirmed that neural network models were used in these studies. This suggests the key role of neural networks in concrete research, which is in entire agreement with the observation in Section \ref{sec:keyword_co-occurrence_analysis}. Another focus of this cluster was durability, such as carbonation \citep{Kwon2010}, chloride penetration \citep{Song2009}, later-age compressive strength \citep{Pala2007}, and residual strength after high temperature \citep{Chang2006}.

The fourth largest cluster with 47 cited references, cluster \#4, was identified as ``load-carrying capacity'', which is not a clear labeling. We reviewed the content of the six publications in this cluster and identified shear capacity \citep{Stegemann2002} or elastic modulus \citep{Gesoglu2010,Demir2008,Yan2010} as the focus. Cluster \#5 labeled as ``cement paste'' had 46 cited references but only two citing articles \citep{Stegemann2002,Stegemann2001}. These two studies investigated the effects of pure inorganic or toxic metal compounds on compressive strength \citep{Stegemann2002} and setting time \citep{Stegemann2001} of cement paste systems by neural network models, providing financial benefits and environmental-friendly solutions for the treatment of industrial wastes by cement-based solidification. Hence, the focus of this cluster could be cement paste or waste solidification. Compared with other clusters, cluster \#5 was the smallest cluster, which implies relatively little attention has been devoted to the research topic so far. 

Note that all the high silhouette values in Table \ref{tab:document_co-citation} were larger than the threshold 0.7, suggesting that each cluster is highly isolated and disjointed from each other with weak links in between \citep{Chen2014} (see Fig.~\ref{fig:document_co-citation_timeline}). This highlights the fact that researchers do not cite relevant work outside their clusters \citep{Structures}, and thus fail to draw on a wide range of knowledge sources \citep{Darko2020,Nerur2008}. Consequently, the current body of research on AI in the concrete domain appears inward-looking, not benefiting from applicable theories or concepts from other research domains \citep{Hosseini2018}. As AI itself is a concept from the field of computer science, it is expected that interdisciplinary research could expand the frontier of knowledge in the concrete field.

\section{Discussion and future trends}
\noindent
AI techniques have entered the concrete science toolbox in the early 1990s and an exponential increase in relevant publications has been witnessed from the last decade (Fig.~\ref{fig:publication_year}). This trend confirms the growing research interest of AI in the field of concrete materials. Existing literature (period: 1990--2020) on AI in concrete materials was examined in the preceding sections by scientometric tools, including keyword co-occurrence analysis (Section \ref{sec:keyword_co-occurrence_analysis}) and documentation co-citation analysis (Section \ref{sec:document_co-citation_analysis}). The scientometric analysis extends earlier review studies of the field by providing a quantitative interpretation of the past and present knowledge structure and offering a complete picture and understanding of AI literature in the concrete domain. Most research efforts have been directed towards prediction of concrete properties, in particular compressive strength, and substantial successes have been achieved by the use of neural network models. For example, the highest cited work by Yeh \citeyearpar{Yeh1998} (Table \ref{tab:top10_most_cited}) employed neural network models on a dataset of over 1,000 samples to predict the compressive strength of high performance concrete, and obtained a high coefficient of determination ($R^2$) of about 0.91 to 0.92. These neural network models with concrete mixture constituents as the input significantly outperformed the traditional regression analysis merely based on water-to-cement ratio and age ($R^2 \approx 0.57$). Since then, the neural network approach has received the most attention for predicting properties of different types of concrete \citep{Young2019,Duan2013,Ramkumar2020,Dingqiang2020,Dias2001,BenChaabene2020}. In this section, we provide our perspectives on the key challenges and opportunities within the domain of AI in concrete science based on the results presented earlier.

\subsection{Data} \label{sec:data}
\noindent
As discussed in Section \ref{sec:keyword_co-occurrence_analysis}, existing research efforts have focused on the prediction of hardened properties of concrete materials, while modeling behaviors of concrete in fresh state has received far less attention in current literature. This could be attributed to the difficulty in constructing datasets for concrete fresh properties given that measurement principles and results vary among testing methods or instruments (e.g., rheometers). Further examination of the retrieved 389 publications showed that the majority of studies have relied heavily on small datasets with less than 200 experimental data points \citep{Li2022a,BenChaabene2020}, which could be insufficient for AI techniques, especially neural network models with a large number of parameters to be calibrated, and may lead to overfitting of AI models and poor generalization performance on new datasets. 

The lack of large and universal datasets for concrete materials becomes a major challenge towards the wider adoption of AI in the construction industry. Many tens of thousands of concrete mixtures as well as their physical properties have been reported in the literature, but collecting these data could be a daunting data mining problem. Research articles are written in different formats by researchers in different laboratories, using different experimental parameters; methods and results are presented by different terms with inconsistent levels of completeness; thus, manual extraction of data from literature could be rather time-consuming and labor-intensive \citep{Cao2018a}. In addition, null or negative results tend to be discarded and not published, although these results would be extremely useful in the context of AI. These lead to a call for standardization of data reporting criteria and sharing of research data, which are a big step towards open science \citep{Gewin2016}.

There are also other approaches worth exploring when dealing with data sparsity issues. One promising yet understudied topic in concrete research is transfer learning, where the knowledge learned from one material or property can be used to develop models for a similar material or property \citep{Pan2010,Ravinder2021}. As discussed in Section \ref{sec:keyword_co-occurrence_analysis}, existing literature has mainly focused on specific concrete types (e.g., high performance concrete) and concrete properties (e.g., compressive strength). Thus, it is reasonable to assume that more datasets related to these types of concrete and properties are available. By using transfer learning, reliable models could be developed for other types of concrete or other properties that have received far less attention with only sparse data available \citep{Gopalakrishnan2017,Yang2020,Moussa2020,Tong2020}. 

Another solution to data sparsity is extracting composition-property datasets via natural language processing or text mining approaches \citep{Batra2020}. Data are typically presented in an unstructured form in a research article: researchers may report the compositions of concrete mixtures in one table, the testing conditions and experimental parameters in the body text of the methods section, and then the final properties in figures within the results section \citep{Olivetti2020}. Given the highly distributed nature of data, automated mining information from the literature could be more effective and reliable than manual collection, although there has been no work attempting to apply this approach in the concrete domain.

\subsection{Models}
\noindent
Neural network models have been extensively used in current concrete literature (see Sections \ref{sec:keyword_co-occurrence_analysis} and \ref{sec:document_co-citation_analysis}). The success of neural network models in concrete research provides a glimpse into possible future applications. An emerging trend is the development and deployment of convolutional neural networks, a popular class of deep neural networks or deep learning \citep{Krizhevsky2012,Martinez2019,Khallaf2021}. Deep networks herein refer to neural networks with deeper architectures and more processing layers. Table \ref{tab:top10_most_cited} presents two research articles that were listed among the top cited publications and highly related to the applications of convolutional neural networks \citep{Gopalakrishnan2017,Dorafshan2018}. These two studies have demonstrated the advantages of using convolutional neural networks for crack detection in field practice, e.g., in concrete bridges \citep{Dorafshan2018} and pavements \citep{Gopalakrishnan2017}. This can be attributed to the high accuracy of convolutional neural networks to process data with grid patterns, e.g., images \citep{Krizhevsky2012,Yamashita2018}. While these techniques have been explored for monitoring concrete structures at macro scales, there have been limited studies on their applications in concrete microstructural analyses (e.g., scanning electron microscopy and X-ray computed tomography) \citep{Tong2020,Dong2020,Lorenzoni2020,Song2020}, where a large number of images could be produced and require extensive efforts from researchers to manage and interpret.

Many AI models have been treated as ``black boxes'' in concrete research, especially for neural networks that often provide accurate predictions at the cost of high model complexity. One challenge that hinders their systematic adoption in the construction industry is the lack of interpretability in AI models \citep{Naser2021}. Limited research efforts have been devoted to unravel how or why a model predicts the way it does (i.e., the cause-and-effect relationship). Besides building an accurate model, it is crucial to gain insights and knowledge from the model in the concrete domain. Testable hypotheses resulting from interpretable models can be cross-validated by proposed experiments. Interpretability has been an active area of research in computer science \citep{Lipton2018,Montavon2017,Molnar2019,Molnar2020}. Moving forward, AI models that offer scientific understanding will become increasingly desirable in the concrete research field.

A deep dive into the retrieved concrete literature reveals that most AI research has been based on commonly accessible AI models with a main emphasis on parameter tuning. As these models are designed for data mining purposes, they require large datasets and depend to a large degree on the information contained in the data without considering known scientific principles \citep{Karpatne2017a}. This raises challenges due to the inadequacy of the available data in concrete science (discussed in Section \ref{sec:data}). Recent advances in physics-guided data science have illustrated the potential of integrating known physical laws or scientific knowledge (e.g., the inverse relationship between compressive strength and water-to-cement ratio in the context of concrete research) with AI models \citep{Willard2020,Karpatne2017}. Such an approach offers novel solutions to increase model interpretability, enhance generalization performance, and reduce the amount of data needed for training.

%The preceding sections have demonstrated an intensification of research activities on AI within the concrete domain over the past decades. The scientometric analyses, including keyword co-occurrence analysis (Section \ref{sec:keyword_co-occurrence_analysis}) and documentation co-citation analysis (Section \ref{sec:document_co-citation_analysis}), have identified main research interests and citation patterns in this area, offering a complete picture and understanding of the current knowledge body of AI literature in the concrete field. Results revealed that 

\section{Conclusion}
\noindent
AI has already been utilized in concrete research for almost three decades, where it has advanced the numerical modeling and prediction of concrete properties. This chapter has examined the existing literature using scientometric approaches such as keyword co-occurrence analysis and document co-citation analysis. Results revealed that current research interests have focused on the application of neural network models in concrete compressive strength prediction, while other AI techniques as well as other physical measures remain understudied. This implies that the existing research on AI in concrete science is still immature, with the power of their merging far from being fully realized. With continued advances in data (e.g., reporting standardization, data sharing, knowledge transferring by transfer learning, and automated data collection by natural language processing or text mining) and models (e.g., deeper architectures, higher interpretability, and more physics-aware), a veritable intelligence ecosystem is emerging in concrete science and will have a profound impact on this field.

\section*{Acknowledgements}
\noindent
The authors would like to thank Dr.~Ismaila Dabo, Dr.~Jinyoung Yoon, and Rui Zhang for their insightful thoughts and valuable discussions that form the basis of this work.

\bibliographystyle{elsarticle-harv}
\biboptions{sort&compress}
\bibliography{ML.bib}

\end{document}